\documentclass[runningheads]{llncs}
\usepackage{graphicx}
\usepackage{amsmath}
\usepackage{amssymb}
\usepackage{booktabs}
\usepackage{multirow}
\usepackage{hyperref}

\usepackage{pifont}
\begin{document}

\renewcommand{\thefootnote}{}

\title{MGTR: End-to-End Mutual Gaze Detection with Transformer}\footnotetext{This work is supported in part by National Key Research and Development Project under Grant 2019YFB1310604, in part by National Natural Science Foundation of China under Grant 62173189.}
%
%
\author{Hang Guo \and Zhengxi Hu \and Jingtai Liu \thanks{Corresponding author}} 

\authorrunning{Guo et al.}
%
\institute{Nankai University, Tianjin, China \\ \email{\{1911610, hzx\}@mail.nankai.edu.cn, liujt@nankai.edu.cn}}

\maketitle              

\begin{abstract}
People's looking at each other or mutual gaze is ubiquitous in our daily interactions, and detecting mutual gaze is of great significance for understanding human social scenes. Current mutual gaze detection methods focus on two-stage methods, whose inference speed is limited by the two-stage pipeline and the performance in the second stage is affected by the first one. In this paper, we propose a novel one-stage mutual gaze detection framework called Mutual Gaze TRansformer or MGTR to perform mutual gaze detection in an end-to-end manner. By designing mutual gaze instance triples, MGTR can detect each human head bounding box and simultaneously infer mutual gaze relationship based on global image information, which streamlines the whole process with simplicity. Experimental results on two mutual gaze datasets show that our method is able to accelerate mutual gaze detection process without losing performance. Ablation study shows that different components of MGTR can capture different levels of semantic information in images. Code is available at \href{https://github.com/Gmbition/MGTR}{https://github.com/Gmbition/MGTR}

\keywords{End-to-End Mutual Gaze Detection \and One-stage Method \and Mutual Gaze Instance Match}
\end{abstract}
%
%
\section{Introduction}

Containing rich information, the gaze plays an important role in reflecting the attention,
intention and emotion of one person \cite{admoni2017social,abele1986functions}. Among all kinds of gaze, mutual gaze is indispensable in building the bridge between two minds \cite{argyle1976gaze,loeb1972mutual}. From mutual gaze, one can infer the willingness to interact and the strength of the relationship. Moreover, mutual gaze can also be used for people connection analysis in social scene interpretation and is an important clue for Human-Robot-Interaction. For these reasons, it is very promising to achieve automatic mutual gaze detection.

\begin{figure}[t]
\centering
\includegraphics[width=0.97\linewidth]{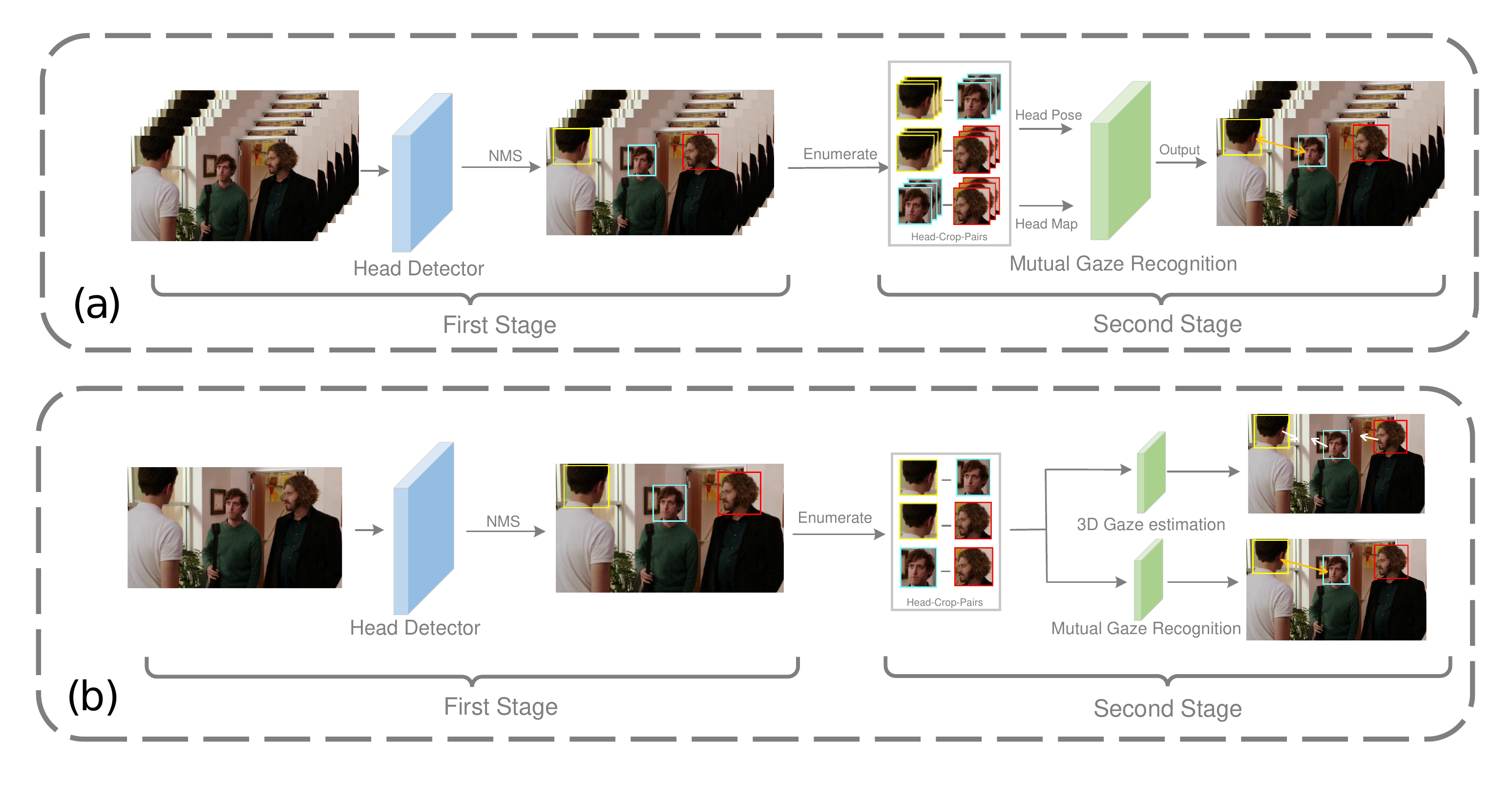}
\caption{An illustration of two-stage methods. The method in (a) uses a pretrained head detector and performs mutual gaze recognition by exploiting pose and position information of enumerated head-crop-pairs. The method in (b) also uses a head detector and utilizes pseudo 3D gaze to boost mutual gaze detection. It can be seen that methods in (a) and (b) both perform head detection in the first stage and need to enumerate all head-crop-pairs in one image which slows down the inference process.}
\label{fig:1}
\end{figure}

The target for end-to-end mutual gaze detection is to detect all the human heads in the scene and then recognize whether any two people are looking at each other. Previous studies \cite{marin2019laeo,doosti2021boosting,marin21pami} have got favorable results by dividing the whole process into two stages: detect all human heads in the scene and then regard the recognition of mutual gaze as a binary classification problem. 
Specifically, Marin \textit{et al} proposed a video based method \cite{marin21pami} (Fig.\ref{fig:1}(a)) that first detects all human heads through a pretrained head detector and then enumerates all head-crop-pairs and put them into a classification network to identify whether two people are looking at each other. The image based mutual gaze detection work in Doosti \textit{et al} \cite{doosti2021boosting} (Fig.\ref{fig:1}(b)) uses pseudo 3D gaze to boost mutual gaze detection and also adapts a two-stage strategy.

Although these two-stage methods have achieved promising results, their designs have some shortcomings. Firstly, the classifier in the second stage makes inferences based on the local information of the head crop instead of the image global information. For example, body posture is also an important clue for judging mutual gaze. Moreover, the performance of the classification results in the second stage depends on the localization accuracy of the first stage. Furthermore, when there are many people in the scene, the computational cost will also increase due to the need to enumerate all detected heads which will slow down the inference process.

\begin{figure}[t]
\centering
\includegraphics[width=0.97\linewidth]{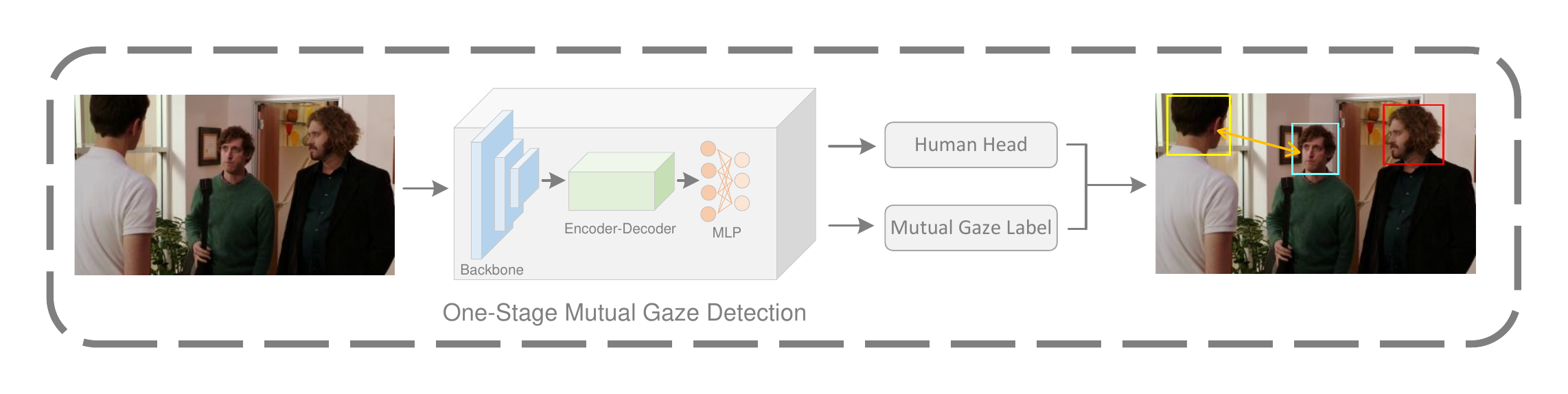}
\caption{An illustration of proposed one-stage method.
Utilizing the designed mutual gaze instance triples, our proposed one-stage method can detect all human heads and the corresponding mutual gaze labels in parallel, which accelerates the pipeline of mutual gaze detection.}
\label{fig:1.1}
\end{figure}

To overcome these limitations, inspired by the attention mechanism in Transformer Network \cite{vaswani2017attention}, we propose a \textit{one-stage} mutual gaze detection model called Mutual Gaze TRansformer or MGTR (Fig.\ref{fig:1.1}) which can detect all human heads in the scene and simultaneously identify whether there is a mutual gaze based on the global image information. By designing mutual gaze instance triples, we improve the mutual gaze detection process from serial to parallel, which greatly accelerates the inference speed without losing performance.

Our proposed MGTR consists of four following modules: A Backbone for feature extraction, Transformer Encoder, Transformer Decoder, and a Fully Connected Neural Network for mutual gaze instance prediction. First, a convolutional neural network is used to extract image features, then a one-dimensional vector is generated by flattening the feature map and we combine it with the positional encoding \cite{parmar2018image,bello2019attention} to get the input of Encoder. After that, the output of Encoder combined with learnable mutual gaze queries are passed through Decoder to model connections between different people, and finally, the encoded mutual gaze queries are passed through a fully connected neural network to output the mutual gaze instance as the result of our model.

Overall, our main contributions are as follows:
\begin{itemize}
    \item We build  a \textit{one-stage} model called MGTR which combines the human head detection and the mutual gaze recognition. To the best of our knowledge, this is the first work that integrates mutual gaze detection task into a one-stage method.
    
    \item Modeling the location information of people and the relationship between them using global information instead of head image crops.
    
    \item Our model outperforms the state-of-the-art method on end-to-end mutual gaze detection task. Moreover, MGTR can perform faster mutual gaze detection.
\end{itemize}

\section{Related Work}

In this section, we first review methods for gaze estimation which encompass a variety of gaze types (Section \ref{section 2.1}). Then go down to the literature of mutual gaze detection (Section \ref{section 2.2}). 
At last one-stage detection methods (Section \ref{section 2.3}) will be reviewed.

\subsection{Gaze Estimation in Social Scenarios}
\label{section 2.1}
Eye gaze can convey rich information and is closely related to the attention, intention, and emotion of a person, even people from different cultures may share a similar meaning of eye gaze \cite{kleinke1986gaze}. Recently, in the computer vision community, there are also a lot of research focusing on social scenario gaze estimation and yielding promising results. For example, Lian \textit{et al} \cite{lian2018believe} proposed a solution for gaze point prediction of the target persons. Fan \textit{et al} \cite{fan2018inferring} proposed a spatial-temporal modeling method to detect people's looking at the same target simultaneously. Zhuang \textit{et al} proposed MUGGLE \cite{zhuang2019muggle}, an approach that is suitable for massive people's shared-gazing. In order to detect whether two people in the video are looking at each other, Marin \textit{et al} \cite{marin2019laeo,marin21pami} proposed a method based on the spatial and temporal information to solve this problem. To understand different types of eye gaze in a group of people, Fan \textit{et al} \cite{fan2019understanding} proposed a method to detect multiple types of human gaze, such as single gaze, shared gaze, etc. In this work, we focus on image based one-stage mutual gaze detection.

\subsection{Mutual Gaze Detection}
\label{section 2.2}
Mutual gaze is one of the most common types of social scenarios gaze communication and there are also methods trying to achieve automatic mutual gaze detection. These methods are all two-stage methods consisting of a human head detector in the first stage and a mutual gaze classifier in the second stage. Specifically, Marin \textit{et al} proposed viode-based LAEO-Net \cite{marin2019laeo} and get a promising result in mutual gaze detection by considering both temporal and spatial information. After that, they further modified LAEO-Net to get LAEO-Net++ \cite{marin21pami}, which achieved better performance. Moreover, the image-based approach proposed by Doosti \textit{et al} \cite{doosti2021boosting} takes advantage of multi-task learning by using pseudo 3D gaze to boost mutual gaze detection. However, these methods suffer from the lack of global image information and slow inference speed due to the sequential two-stage architecture.

\subsection{One-stage Detection Method}
\label{section 2.3}
Recently, in the field of computer vision, many research designs have followed a one-stage idea to speed up the processing pipeline. For example, in the field of object detection, the favorable results of SSD \cite{liu2016ssd}, YOLO \cite{redmon2016you}, RetinaNet \cite{lin2017focal}, DETR \cite{carion2020end} and other methods have demonstrated the advantages of one-stage detection methods. Compared with two-stage detection methods, one-stage methods can perform the detection and classification tasks by using only one network and the pipeline is generally simpler, faster, and more computationally efficient so that it is easier to adopt for real-world applications.

\section{Method}

The task of one-stage mutual gaze detection is to give an image and then detect all mutual maze instances in the image in an end-to-end way. In this section, we will first describe the Representation of Mutual Gaze Instances (Section \ref{section:3.1}) followed by the detailed Model Architecture (Section \ref{section:3.2}), after that we will introduce the Strategy for Mutual Gaze Instances Match (Section \ref{section:3.3}) and at last the Loss Function Setting (Section \ref{section:3.4}).

\begin{figure}[t]
\centering
\includegraphics[width=12cm]{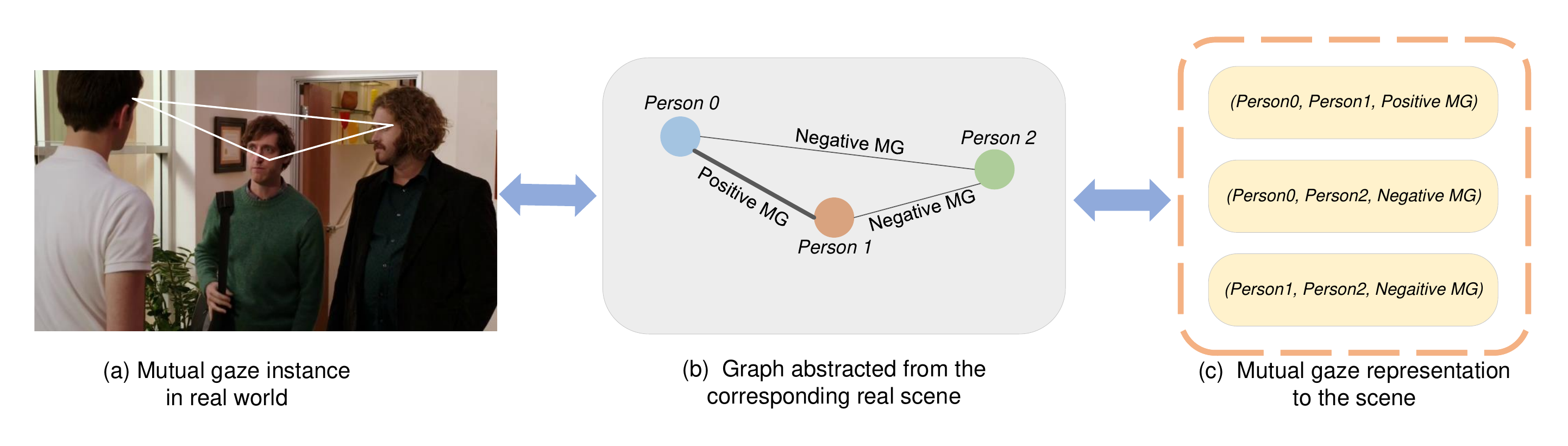}
\caption{Representation of Mutual Gaze Instances in the scene. In the social scene, each individual is uniquely numbered and the enumerated head pairs are order-independent. \textit{MG} means Mutual Gaze.
}
\label{fig:2}
\end{figure}

\subsection{Representation of Mutual Gaze Instances}
\label{section:3.1}
We define a mutual gaze instance as a triple, namely (Person1, Person2, Mutual Gaze Label), where  Person1 and Person2 contain the bounding box coordinates and the class confidence of a person, and Mutual Gaze Label is one when Person1 and Person2 are looking at each other otherwise zero. It is noteworthy that under the mutual gaze detection task, it seems useless to predict the class of each box, however, this setting can be used as a detection threshold when we conduct the test phase in which we need the person class confidence of each box. A detailed example of representing a real scene with a mutual gaze instance is given in Fig.\ref{fig:2}.
Additionally, a mutual gaze instance is unordered, that is to say, the relationship between $i$-th Person and $j$-th Person only needs to be recorded once as (Person$i$, Person$j$, Mutual Gaze label between $i$ and $j$).

\subsection{Model Architecture}
\label{section:3.2}
\begin{figure}[t]
\centering
\includegraphics[width=12cm]{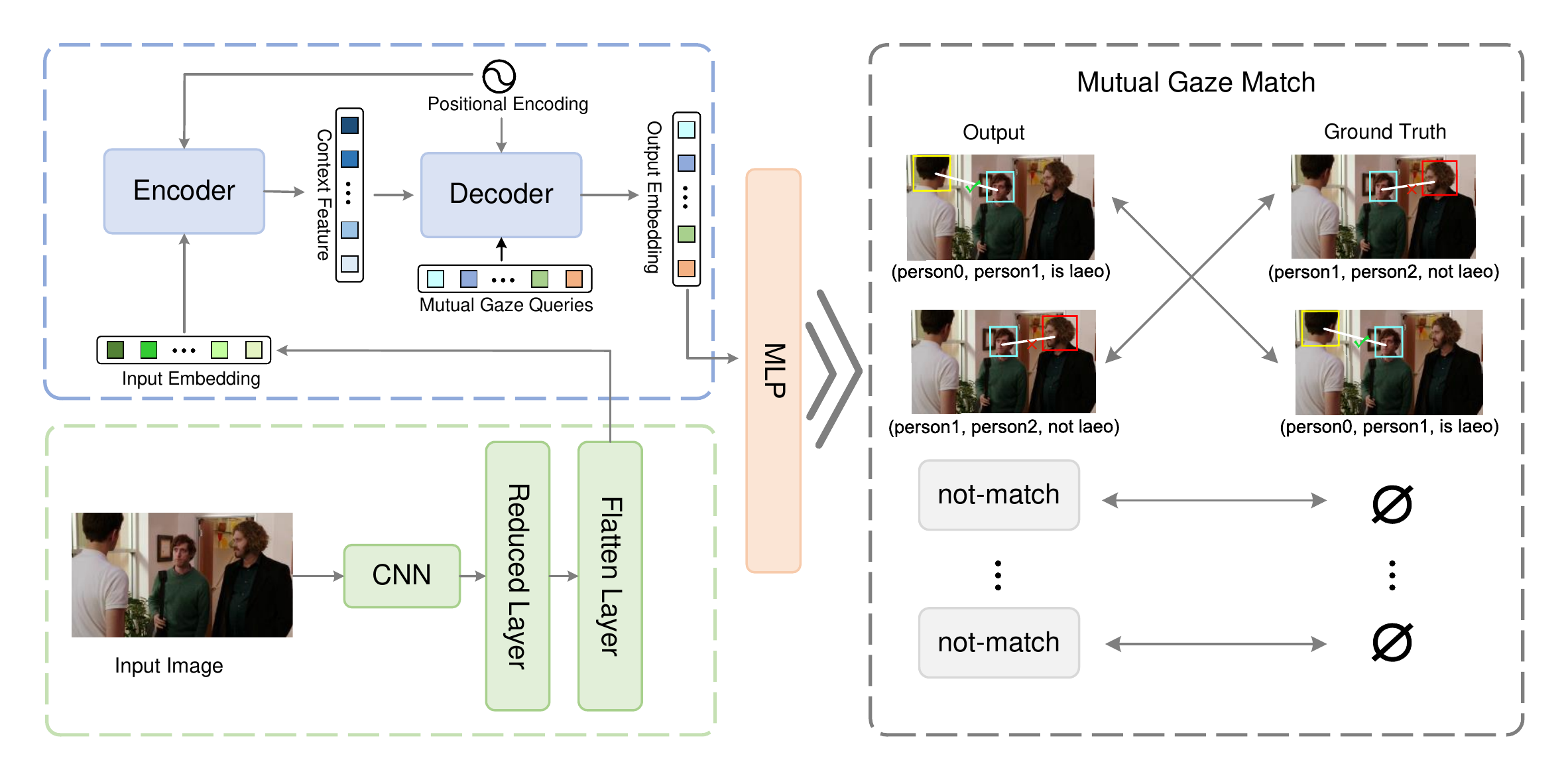}
\caption{An overview of MGTR architecture. It consists of four components: a Backbone to convert input image into an one-dimensional input embedding, an Encoder followed by a Decoder to get the encoded Mutual Gaze Queries, and MLP to predict mutual gaze instances. \textit{laeo} means Looking At Each Other. See Section \ref{section:3.2} for more details.}
\label{fig:3}
\end{figure}

Our proposed MGTR mainly consists of four parts: a Backbone, an Encoder module, a Decoder module, and MLP. Fig.\ref{fig:3} shows an overview of MGTR architecture.

\subsubsection{Backbone}

A convolutional neural network is used to extract features from an input image of original size $[H,W,3]$. After the convolutional neural network, we get a feature map of size $[C, H , W]$. We then reduce the channel dimension of the feature map from $C$ to \textit{d} by a $1\times1$ convolution kernel resulting the new feature map of size $[d, H, W]$. Since Transformer Encoder requires a sequence as input data, we compress the last two dimensions of the new feature map to obtain a flatten feature called \textit{input embedding} of size$[d, HW]$.

\subsubsection{Encoder}
The Encoder layer in MGTR is the same as the standard Transformer Encoder layer, including a multi-head self-attention layer and a feed forward network(FFN). Due to the permutation invariance of the Transformer, we add the \textit{positional encoding} to the input embedding to obtain the Query and Key of Encoder and only use the input embedding as the Value. For the convenience of description, we represent the output of the Encoder as the \textit{context feature}.

\subsubsection{Decoder}
The Decoder layer in MGTR is also the same as the standard Transformer Decoder layer which contains two multi-head attention layers and a feed-forward network. We refer to the $N$ learnable positional embeddings as \textit{mutual gaze queries}. In the multi-head self-attention layer, the Query, Key, and Value all come from either the mutual gaze queries or the sum of the previous decoder layer’s output and mutual gaze queries. As for the encoder-decoder cross attention layer, the Value comes from the context feature generated from the Encoder, the Key is the sum of context feature and positional encoding and the Query is the sum of the multi-head self-attention layer’s output and mutual gaze queries. We denote the output of the Decoder as \textit{output embedding}.  

The self-attention mechanism in the Encoder and Decoder can help us model the positions of different people in the image and the relationship between them. The $N$ output embeddings encoded from $N$ mutual gaze queries are then converted into a mutual gaze instance by the subsequent MLP so that we get $N$ final mutual gaze instance results and we will discuss this part next. 

\subsubsection{MLP for Mutual Gaze Prediction}

After passing the $N$ mutual gaze queries through the Decoder, we get $N$ output embedding, which contains information about the position of the head bounding boxes and the relationship between different people in the image. Then we pass the output embedding into the MLP to predict the mutual gaze instance. Specifically, three one-layer fully-connected neural networks are used to predict the confidence score of Person1, Person2, and Mutual Gaze Label respectively, and two three-layer fully-connected neural networks are used to predict the head bounding boxes of Person1 and Person2. For the human confidence score prediction branch, there are three classes: whether it is a person and \textit{not-match} (the meaning of not-match will be described later). For the mutual gaze confidence prediction branch, there are also three categories that indicate whether Person1 and Person2 are looking at each other and not-match. We then apply Softmax to the results of all confidence prediction branches to obtain normalized confidence. For the branch of head bounding boxes regression, the output dimension of the MLP is four, which represent the normalized center coordinates, width, and height of the bounding boxes respectively.

\begin{figure}[t]
\centering
\includegraphics[width=12cm]{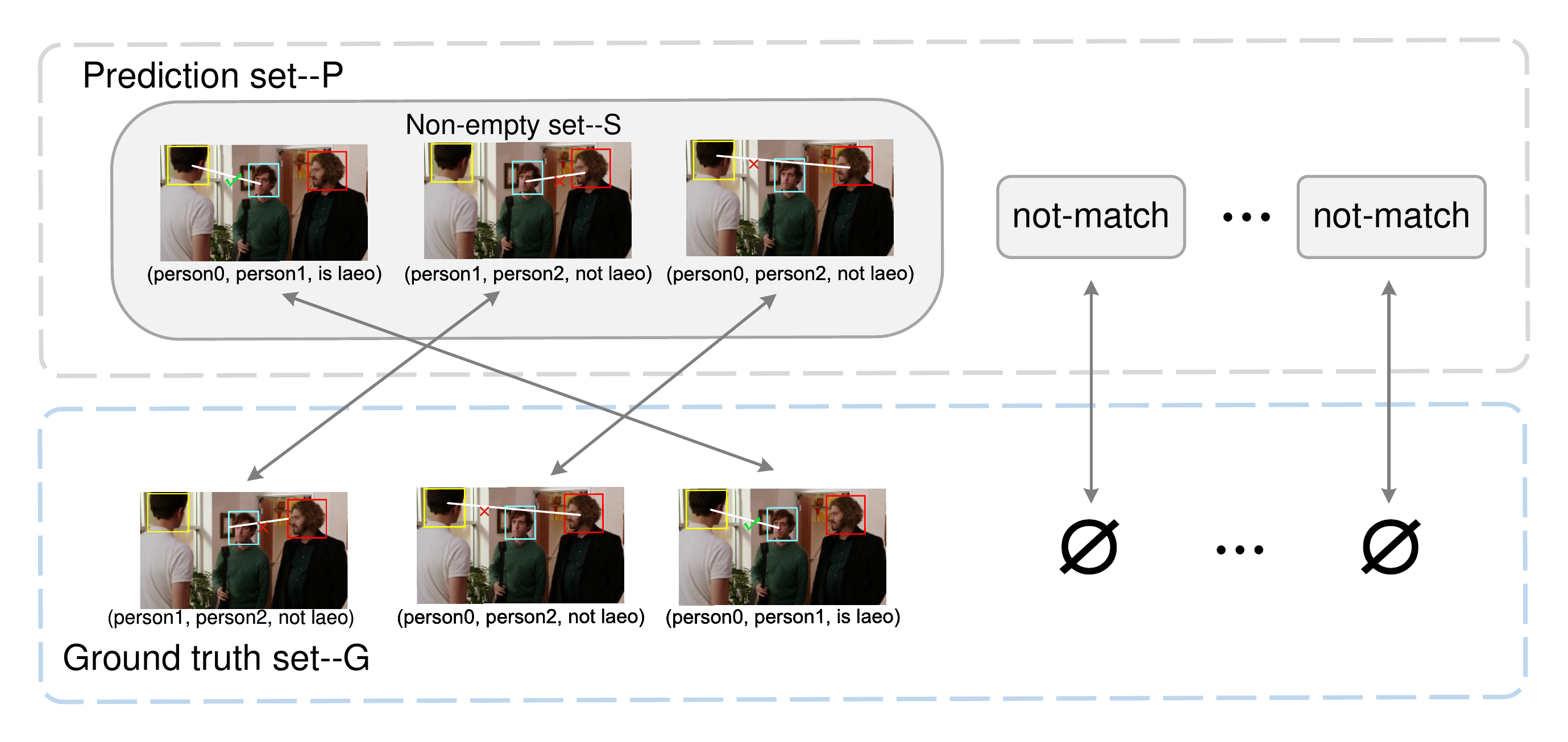}
\caption{An example explaining the mutual gaze instances match. By padding the set $\mathbf{G}$ with $\varnothing$, we make $\mathbf{P}$ and $\mathbf{G}$ the same size, which can be transformed into a bipartite graph matching problem. \textit{laeo} means Looking At Each Other}
\label{fig:4}
\end{figure}

\subsection{Strategy for Mutual Gaze Instances Match}
\label{section:3.3}
After passing the output embedding through MLP, we get $N$ predicted mutual gaze instances. However, the number of ground truth mutual gaze instances is not necessarily $N$ (often less than $N$). So it requires some predicted mutual gaze instances representing empty which we denoted as \textit{no-match}, indicating that these mutual gaze instances do not match any ground truth. To be precise,
we denote the set of ground truth instances as $\mathbf{G}$, and the size of $\mathbf{G}$ is $M$, the set of predicted instances is $\mathbf{P}$ and the size of $\mathbf{P}$ is $N$, then a satisfactory model should output a set which contains $N-M$ instances representing \textit{not-match}.

After solving the problem that the number of ground truth instances and predicted instances are not always equal by designing the \textit{not-match} class, the next key problem is how to match the predicted instances with the ground truth instances. Specifically, we denote the set of elements predicted to be non-empty in $\mathbf{P}$ as $\mathbf{S}$, then the problem we have to solve is how to build a map $\sigma$ from $\mathbf{S}$ to $\mathbf{G}$. It is worth mentioning that the size of $\mathbf{S}$ is not necessarily equal to the size of $\mathbf{G}$. We solve this matching problem in another way: by padding the ground truth set $\mathbf{G}$ with $\varnothing$, we make $\mathbf{G}$ and $\mathbf{P}$ equal in size. So the above matching problem is transformed into a one-to-one bipartite matching problem between $\mathbf{P}$ and $\mathbf{G}$. In this work, we use the Hungarian algorithm \cite{kuhn1955hungarian} to solve this problem. A more concrete example can be seen in Fig.\ref{fig:4}.

\subsection{Loss Function Setting}
\label{section:3.4}
Assume the mapping from the predicted set $\mathbf{P}$ to the ground-truth set $\mathbf{G}$ is denoted as $\sigma(i)$, which means the $i$-th element in the set $\mathbf{P}$ will be mapped to the $\sigma(i)$-th element in $\mathbf{G}$. We design the matching cost of the Hungarian algorithm as follows.

\begin{equation}
    \mathcal{L}_{match} =\sum_{i=1}^{N}[\beta_1 \mathcal{L}_{class}(c_i,c_{\sigma(i)}) +\beta_2 \mathcal{L}_{box}(b_i,b_{ \sigma(i)})]
    \label{equ:1}
\end{equation}

where $\mathcal{L}_{class}(c_i,c_{\sigma(i)})$ represents the cost between the $i$-th mutual gaze instance from $\mathbf{P}$ and the $\sigma(i)$-th ground truth from $\mathbf{G}$ in human class confidence and mutual gaze confidence, and we call it the \textit{class loss function}. $\mathcal{L}_{box}(b_i,b_{\sigma(i)})$ represents the cost between the $i$-th mutual gaze instance from $\mathbf{P}$ and the $\sigma(i)$-th from $\mathbf{G}$ in head bounding boxes regression, and we call it the \textit{head bounding box regression loss function}. $\beta_1$ and $\beta_2$ are hyperparameters used to measure the weight between these two types of losses. The specific forms of these two losses are discussed below.

For the class loss function, we denote the value of mutual gaze confidence as $p_{i}^{gaze}$ and use $p_{i}^{h_1}$ and $ p_{i}^{h_2}$ to represent the human class confidence. Under this representation, the class loss function is defined as follow.

\begin{equation}
    \mathcal{L}_{class}(c_i,c_{\sigma(i)}) =\alpha_1 p_{i}^{h_1}+\alpha_2 p_{i}^{h_2}+\alpha_3 p_{i}^{gaze}
    \label{equ:2}
\end{equation}

For the head bounding box regression loss function, the definition is as follow.
\begin{equation}
    \mathcal{L}_{box}(b_i,b_{\sigma(i)})=\gamma_1
    \ell_1(b_i,b_{\sigma(i)})+\gamma_2 \text{GIoU}(b_i,b_{\sigma(i)})
    \label{equ:3}
\end{equation}

where $\ell_1(\cdot)$ is the $L_1$ loss. We also added the GIoU loss \cite{rezatofighi2019generalized} into the head bounding box regression loss function, and we will confirm the importance of GIoU loss in the later Ablation Study.

By defining the above costs from the $i$-th element in $\mathbf{P}$ to the $\sigma(i)$-th element in $\mathbf{G}$, we can solve the following optimal bipartite match problem based on the Hungarian algorithm.
\begin{equation}
    \sigma ^* =\mathop{\text{argmin}}\limits_{\sigma} \mathcal{L}_{match}
    \label{equ:4}
\end{equation}

After the optimal bipartite matching problem is solved by the Hungarian algorithm, we can next calculate the loss function for training. In this work, the definition of loss function is almost the same as in Eq.\ref{equ:1}, the difference is that in $\mathcal{L}_{class}$ we use cross-entropy loss instead of negative predicted probability and there are also some subtle differences in the settings of hyperparameters as well (more details can be seen in Section \ref{section:4.4}). 

Different from previous mutual gaze detection methods which first train a head detector, and then freeze the head detector parameters and train the mutual gaze classifier. The method in our work optimizes the class loss function and the head bounding box regression loss function simultaneously during the model training process.

\begin{figure}[t]
\centering
\includegraphics[width=11cm]{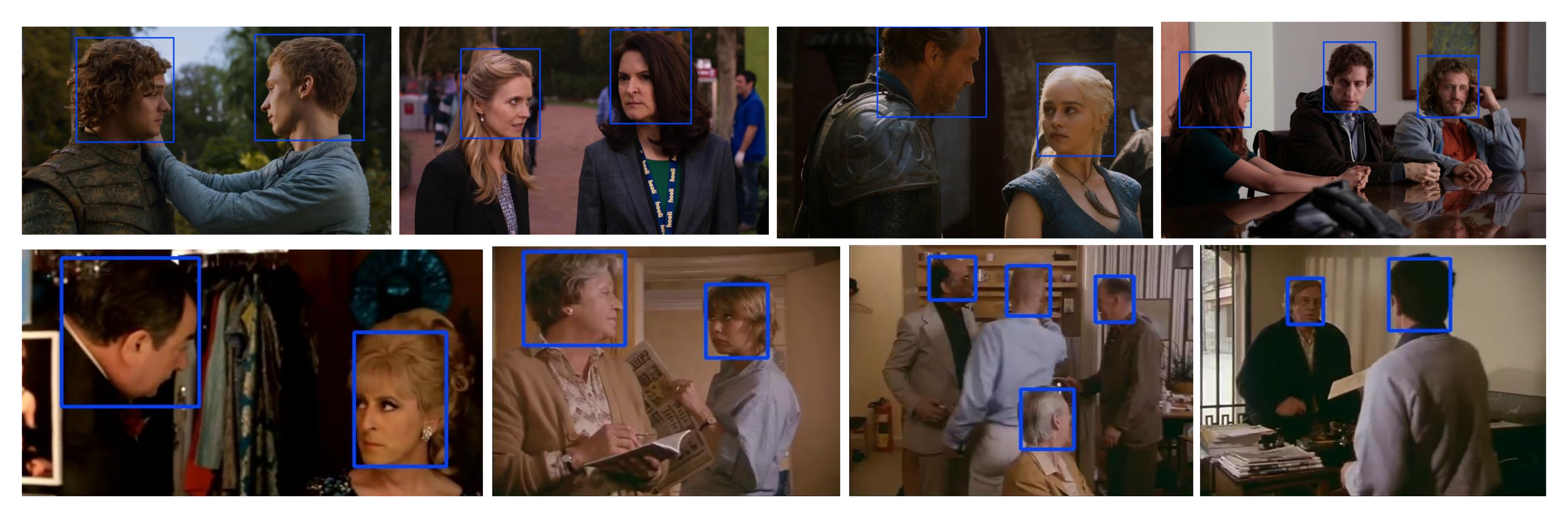}
\caption{Some examples of UCO-LAEO and AVA-LAEO datasets. The images in the first row come from the UCO-LAEO dataset which contains both head bounding boxes and mutual gaze labels. The images in the second row are from the AVA-LAEO dataset which only contains mutual gaze labels and we approximate the ground-truth head bounding box coordinates by using a pre-trained head detector \cite{liu2016ssd}.}
\label{fig:5}
\end{figure}

\section{Experiments}

\subsection{Datasets}

\subsubsection{UCO-LAEO \cite{marin2019laeo}}

This dataset consists of detailed mutual gaze instance annotations with human head bounding boxes and mutual gaze labels. The original dataset has been manually divided into positive and negative instances for sample balance, which means there may be some instances in one image that are not annotated as negative instances. In this work, since we need to detect all the instances in one image, so we only use positive samples from original annotations and treat all remaining instances as negative samples.

\subsubsection{AVA-LAEO \cite{marin2019laeo}}

This dataset has a broad coverage that reaching 50,797 video frames generated from \textit{Atomic Visual Actions} dataset(AVA v2.2) \cite{murray2012ava}. Since there are no manual head bounding box annotations in the original AVA v2.2 dataset, we first use a pre-trained head detector \cite{liu2016ssd} to generate all head bounding box coordinates in each frame as head bounding box ground truth before our training starts. Similarly, we also only use positive instances from the original dataset annotations and regard all the remaining as negative. 

Some examples of UCO-LAEO and AVA-LAEO datasets are shown in Fig.\ref{fig:5}.

\subsection{Evaluation Metric}

In this work, we use the \textit{mean average precision}(mAP) as a measure of how well our model performs on the two datasets. It is worth mentioning that although our task is similar to a binary classification task, we do not use the AP of a single class as an evaluation metric. This is because our task is to simultaneously predict accurate human head positions as well as mutual gaze labels. So we need to consider both classes of AP to include the evaluation of head position localization accuracy. For example, if it is known that the model predicts the mutual gaze label in the current instance to be positive but in fact it is a wrong prediction, it does not necessarily mean that the ground-truth mutual gaze label is negative because it is also possible that the model does not match the head bounding box with ground truth when predicting. So, an instance predicts correctly if and only if it locates the head boxes of the two people in the correct position and predicts correctly whether the two are looking at each other. This criterion requires both the AP in two classes should be high.

\subsection{Current State-of-the-Art Approach}

The current state-of-the-art approach for video-based mutual gaze detection task is LAEO-Net++ proposed by Marin \textit{et al} \cite{marin21pami}. The image-based state-of-the-art method is proposed by Doosti \textit{et al} \cite{doosti2021boosting} and we call it \textit{Pseudo 3D Gaze} in this paper.

The two aforementioned works are different from ours. Specifically, both works above only detect mutual gaze without detecting human head position, while MGTR detects above both whose task is more difficult. Moreover, LAEO-Net++ is a video-based method which uses ground-truth human head box as model input and utilizes temporal connection among neighbor video frames resulting in the input containing more prior information, while ours is an image-based method whose input is only one single image. Since these differences, we modify them accordingly for a fair comparison. Specifically, as for LAEO-Net++, since we focus on image-based mutual gaze detection in this work, so we plan to use its image-based version introduced in \cite{doosti2021boosting}. However, we cannot get the modified LAEO-Net++ for comparison since the code in \cite{doosti2021boosting} is not open source. So we directly use the performance numbers of image-based \textit{LAEO-Net} reported in \cite{doosti2021boosting}. As for Pseudo 3D Gaze which focuses on the performance of the second stage, we add a head detector in front of the original model for end-to-end mutual gaze detection, a detailed description of this model can be seen in Section \ref{section:4.4}.

\subsection{Implementation Details}
\label{section:4.4}

\subsubsection{Data Augmentation}

We normalize the input image by using the mean and std from ImageNet \cite{krizhevsky2017imagenet}. To improve the robustness of the model, we randomly apply horizontal flipping, adjusting brightness and contrast, random cropping, and random resize (to enable the model to detect instances at multiple scales). 

\subsubsection{Hyperparameters Settings} 

In order to balance the class cost and head bounding box regression cost, we set $\beta_1=1.2$, $\beta_2=1.0$ in the Hungarian algorithm matching process and $\beta_1=\beta_2=1.0$ in the training loss function. At the same time, we set $\alpha_1=\alpha_2=1.0$, $\alpha_3=2.0$ both in the Hungarian algorithm and loss function to make our model focus more on judging the existence of mutual gaze. In the head bounding box regression loss function, we set $\gamma_1=5.0$, $\gamma_2 = 2.0$.

\subsubsection{Training Settings}

We used Resnet50 \cite{he2016deep} with frozen batchnorm layer as Backbone for MGTR, the number of Encoder and Decoder layers are both set to 6, the same as in \cite{zou2021end}. Both Backbone and Encoder-Decoder use the pretrained parameters from COCO \cite{lin2014microsoft} pretrained DETR \cite{carion2020end}. The number of mutual gaze queries is set to 100. In the training phase, we choose the batch size to be 8, and we use AdamW \cite{loshchilov2017decoupled} as an optimizer, with a constant learning rate of 1e-4 in Encoder-Decoder and 1e-5 in Backbone, we train the model until the performance on the test set no longer improves.

\subsubsection{Description for Pseudo 3D Gaze using Head Detector}

Since the Pseudo 3D Gaze method in \cite{doosti2021boosting} uses the ground truth head bounding boxes, for a fair comparison, we set this baseline that consists of two parts: a head detector and a mutual gaze classifier. We use the head detector proposed in \cite{zhang2017s3fd} for head box detection. The classifier of this baseline adopts the same network as the one proposed by Doosti \textit{et al} and also uses the pseudo 3D gaze to boost the training process. During training, we only train the mutual gaze classifier with randomly initialized weights by using the ground truth head bounding boxes. During testing, we first detect each head bounding boxes through the pretrained detector and then pass the paired detected head crops through the mutual gaze classifier to get the result for each mutual gaze instance.

\setlength{\tabcolsep}{4pt}
\begin{table}[t]
\centering
\caption{
Comparison with State-of-the-Art Method. $\text{FPS}_\text{{ext}}$ refers to the number of images processed per second in the extreme social scene with more than four people, and $\text{FPS}_\text{{all}}$ refers to the number of pictures processed per second in scenes averaged across the whole test set. The FPS is evaluated using NVIDIA 3090TI GPU. Since we have no access to the code of image based LAEO-Net in \cite{doosti2021boosting}, we do not evaluate the FPS of image based LAEO-Net. w/ GT and w/t GT respectively indicate whether to use ground truth head bounding boxes as model input. The mAP represents positive class's AP in two-stage methods and two classes's average AP in one-stage methods. *Number reported from \cite{doosti2021boosting}.}
\label{table:1}
\renewcommand\arraystretch{1.3}
\resizebox{\textwidth}{!}{%
\begin{tabular}{@{}ccccccccccc@{}}
\toprule[1.5pt]
\toprule
\multicolumn{2}{c}{\multirow{3}{*}{Method}} &
 \multirow{3}{*}{End-to-end training}  &
  \multicolumn{2}{c}{$\text{FPS}_{\text{ext}}$} &
  \multicolumn{2}{c}{$\text{FPS}_{\text{all}}$} &
  \multicolumn{4}{c}{mAP} \\
\multicolumn{2}{c}{} &
   &
  \multirow{2}{*}{UCO} &
  \multirow{2}{*}{AVA} &
  \multirow{2}{*}{UCO} &
  \multirow{2}{*}{AVA} &
  \multicolumn{2}{c}{UCO-LAEO} &
  \multicolumn{2}{c}{AVA-LAEO} \\
\multicolumn{2}{c}{} &
   &
   &
   &
   &
   &
  w/ GT &
  w/t GT &
  w/ GT &
  w/t GT \\ \midrule
\multirow{2}{*}{$Two-stage^*$} &
  Iamge based LAEO-Net &
  \ding{55} &
  - &
  - &
  - &
  - &
  55.9 &
  - &
  70.2 &
  - \\
 &
  Pseudo 3D Gaze &
 \ding{55} &
  0.51 &
  0.93 &
  0.49 &
  0.89 &
  \textbf{65.1} &
  46.7 &
  \textbf{72.2} &
  52.3 \\ \midrule[0.5pt]
$One-stage$ &
  MGTR (ours) &
  \ding{51} &
  \textbf{78.06} &
  \textbf{14.56} &
  \textbf{9.18} &
  \textbf{10.81} &
  - &
  \textbf{64.8} &
  - &
  \textbf{66.2} \\ \bottomrule[1.5pt]
\end{tabular}%
}
\end{table}
\setlength{\tabcolsep}{1.4pt}

\subsection{Comparison with State-of-the-Art Method}

Table \ref{table:1} shows the quantitative results compared with the state-of-the-art method in terms of FPS and mAP on UCO-LAEO and AVA-LAEO datasets.

For UCO-LAEO dataset, MGTR processing is more than 150 times faster than the two-stage approach in extreme social scenes with more than four people and nearly 17 times faster in scenes averaged across the test set. Moreover, MGTR achieves 64.8\% mAP, an 18.1\% increase over the Pseudo 3D Gaze using detected head bounding boxes. Meanwhile, our method is also comparable with image-based SoTA methods who use ground truth head bounding boxes, with only a difference of 0.3\%. The good performance of MGTR on mAP shows that MGTR can handle the imbalance of positive and negative samples in the training set well.

As for AVA-LAEO dataset, MGTR processes each image more than 15 times faster than the two-stage method in more than four people social scenes and more than 12 times in the average scene across the test set. Besides, MGTR gets a 66.2\% mAP score, a 13.9\% increase compared with the baseline using detected head bounding boxes.

\setlength{\tabcolsep}{4pt}
\begin{table}[t]
\begin{center}
\caption{
Ablation Study of MGTR. mAP refers to the average AP in the two classes, $\text{AP}_{\text{rare}} $ refers to the AP of the minority category (usually the positive mutual gaze label), and $\text{AP}_{\text{normal}}$ refers to the majority category, and Recall refers to the average Recall over two classes.
}
\label{table:2}
\renewcommand\arraystretch{1.3}
\resizebox{\textwidth}{!}{
\begin{tabular}{cccccc}
\toprule[1.5pt]
\toprule
Model Setting & \#param & mAP & $\text{AP}_{\text{rare}}$ &$\text{AP}_{\text{normal}}$& Recall \\
\midrule[0.5pt]
NoDataAugmentation   & 41.4M & 53.4 & 52.8 & 54.3 & 68.6 \\
Resnet101Backbone      & 60.3M & 51.3 & 49.7& 53.0 & 65.5\\
NoGIoULoss & 41.4M & 55.8 & 52.8 &58.8 & 67.3\\
DataAug+Resnet50+DIoU&41.4M&55.7&51.1&60.4&69.2\\
DataAug+Resnet50+CIoU&41.4M&60.4&54.5&66.3&68.4\\
\midrule[0.5pt]
Base(DataAug+Resnet50+GIoU)   & 41.4M &  \textbf{64.8} & \textbf{58.3} & \textbf{71.4} & \textbf{75.3} \\
\bottomrule[1.4pt]
\end{tabular}}
\end{center}
\end{table}
\setlength{\tabcolsep}{1.4pt}

\subsection{Ablation Study}
In this part, we design some ablation methods to study how data augmentation, different components of MGTR and loss function setting will affect the performance. We choose the UCO-LAEO dataset and use MGTR with Resnet50 backbone as a base model. Ablation baselines are as follows: (1) No multi-scale and random cropping in data augmentation (2) Using Resnet101 as a backbone, we design this part to study how the complexity of the model will affect performance (3) No GIoU loss, we only use BCE Loss and $L_1$ Loss as loss function. (4) Use DIoU loss \cite{zheng2020distance} instead of original GIoU loss and keep other parts consistent with base model. We design this part to explore the effect of different IoU losses. (5) Use CIoU loss \cite{zheng2020distance} to replace GIoU loss and keep other parts unchanged.

The results of the ablation study are provided in Table \ref{table:2}. It can be seen that data augmentation including multi-scale resize and random cropping are important for training, without which resulting in an 11.4\% drop in mAP. At the same time, the use of Resnet101 as the backbone leads to a decrease on performance, so the more complex the model is not always the better. As can be seen from the fourth-to-last row of Table \ref{table:2}, when the GIoU loss is removed from the loss function, the performance of MGTR on mAP drops by 9.0\%, which indicates that the GIoU loss is indispensable for MGTR to accurately locate each person's head bounding box. When using DIoU loss or CIoU loss to replace the original GIoU loss, the performance on mAP drops by 9.1\% and 4.4\%, respectively, indicating that even though DIoU and CIoU are improvements over GIoU, using GIoU loss still achieves the best performance. This may be due to the fact that both Backbone and Encoder-Decoder in MGTR are initialized using the parameters in DETR pretrained model that also uses GIoU loss. Therefore, using a consistent loss may give better results.

\begin{figure}[t]
\centering
\includegraphics[width=12.5cm]{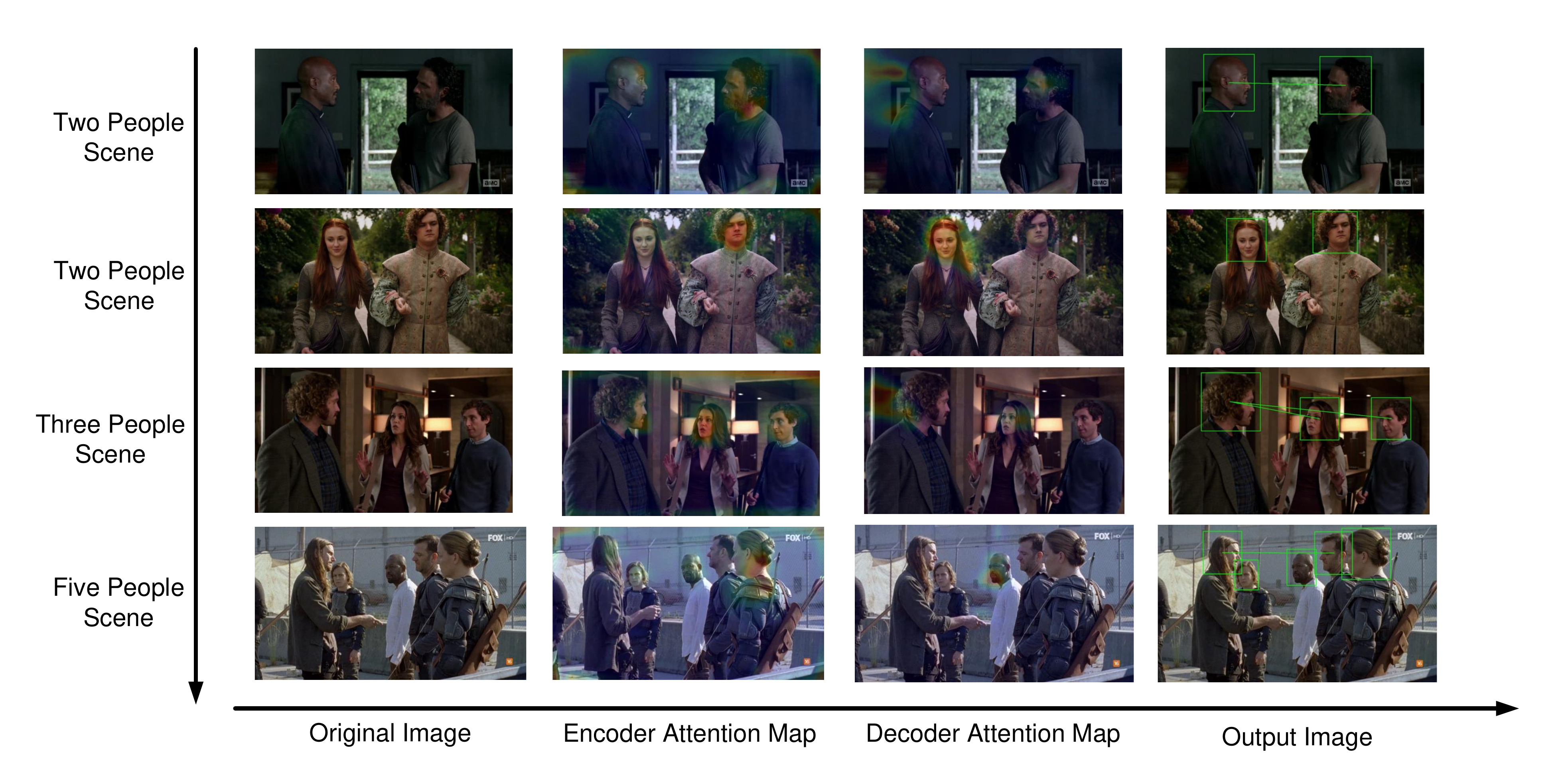}
\caption{
Visualization of last attention layer in Encoder and Decoder and the predicted result by MGTR (from UCO-LAEO dataset). For each row, the first image is the original input image, the second image is the attention weight in Encoder, the third image is the attention weight in Decoder, and the last image is the predicted result by MGTR.}
\label{fig:6}
\end{figure}

\subsection{Qualitative Analysis}

Different from the previous two-stage methods, MGTR simultaneously gives all head bounding boxes and mutual gaze labels in the scene in an end-to-end manner, which does not seem easy to understand. In this part, we will analyze the different roles of Encoder and Decoder of MGTR in different levels of image semantic understanding.

To study the different roles of Encoder and Decoder in MGTR, we visualize the last attention layer of Encoder and Decoder respectively, results can be seen in Fig.\ref{fig:6}. It can be easily seen that the role of  Encoder in MGTR is to find all the head bounding boxes in the social scene, because the head area of each person in the attention-map is given a larger attention weight. After the Encoder finds all the people in the scene, Decoder can find the relationship between different people. In the Decoder's attention-map, we can see that when the mutual gaze label is positive there will be two people's head regions with large attention weights. However, when the label is negative, only one person's head region will be focused. Therefore, we can conclude that the role of Decoder is to predict which pairs of people in the current scene are looking at each other so that model the relationship between different people.

\section{Conclusion}

In this work, we propose a one-stage mutual gaze detection method called Mutual Gaze TRansformer or MGTR to directly predict mutual gaze instances in an end-to-end manner. Different from current two-stage mutual gaze detection methods, MGTR is the first work that integrates human head detection and mutual gaze recognition into one stage which simplifies the detection pipeline. Experiments on two mutual gaze datasets demonstrate that our proposed method can greatly accelerate inference process while improving performance. In the future, we will explore the application of incorporating mutual gaze information into the analysis of the interpersonal relationship, and the detected mutual gaze instance will serve as an important clue for social scene interpretation.

%
%
%
\bibliographystyle{splncs04}
\bibliography{egbib}

\end{document}